\documentclass[a4paper,final,10pt,twocolumn]{elsarticle}
\usepackage[vmargin={20mm,20mm},hmargin={15mm,15mm}]{geometry}

\usepackage[T1]{fontenc}
\usepackage[utf8]{inputenc}
\usepackage{soul} 
\usepackage{natbib}
\usepackage{graphicx}
\usepackage{hyperref}
\usepackage{amssymb}
\usepackage{amsmath}
\usepackage{mathtools}
\usepackage[detect-weight]{siunitx}
\usepackage{multirow}
\usepackage{subcaption}
\usepackage{setspace}
\usepackage[dvipsnames]{xcolor}
\usepackage{threeparttable}

\begin{document}

\title{Joint Scene and Object Tracking for Cost-Effective Augmented Reality Guided Patient Positioning in Radiation Therapy}
\author{Hamid Sarmadi\fnref{fn2}}
\ead{hamid.sarmadi@imibic.org}

\author{Rafael~Mu\~noz-Salinas\corref{cor1}\fnref{fn1,fn2}}
\ead{rmsalinas@uco.es}

 \author{ M.\'Alvaro Berb\'is\fnref{fn3}}
 \ead{a.berbis@htime.org}
 
 \author{Antonio Luna \fnref{fn4}}
 \ead{aluna70@htime.org}
 
\author{R. Medina-Carnicer\fnref{fn1}\fnref{fn1,fn2}}
\ead{rmedina@uco.es}

\cortext[cor1]{Corresponding author}
\fntext[fn1]{Computing and Numerical Analysis Department, Edificio Einstein. Campus de Rabanales, C\'ordoba University, 14071, C\'ordoba, Spain, Tlfn:(+34)957212255}
 \fntext[fn2]{Instituto Maim\'onides de Investigaci\'on en Biomedicina (IMIBIC). Avenida Men\'endez Pidal s/n, 14004, C\'ordoba, Spain, Tlfn:(+34)957213861}
 \fntext[fn3] {HT Médica, Hospital San Juan de Dios.  Avda Brillante 106, 14012, Córdoba, Spain}
\fntext[fn4] {HT Médica,  Clínica las Nieves, Carmelo Torres 2,23007, Ja\'en, Spain}

\begin{frontmatter}
\begin{abstract}
    \textbf{Background and Objective}\\
    The research done in the field of Augmented Reality (AR) for patient positioning in radiation therapy is scarce. We propose an efficient and cost-effective algorithm for tracking the scene and the patient to interactively assist the patient's positioning process by providing visual feedback to the operator. Up to our knowledge, this is the first framework that can be employed for mobile interactive AR to guide patient positioning.    \\ \\
    \textbf{Methods}\\
    We propose a point cloud processing method that combined with a fiducial marker-mapper algorithm and the generalized ICP algorithm tracks the patient and the camera precisely and efficiently only using the CPU unit. The alignment between the 3D reference model and body marker map is calculated employing an efficient body reconstruction algorithm.
    \\ \\
    \textbf{Results}\\
    Our quantitative evaluation shows that the proposed method achieves a translational and rotational error of \SI{4.17}{\milli\meter}/\SI{0.82}{\degree} at 9 fps. Furthermore, the qualitative results demonstrate the usefulness of our algorithm in patient positioning on different human subjects. 
    \\ \\
    \textbf{Conclusion}\\
    Since our algorithm achieves a relatively high frame rate and accuracy  employing a regular laptop (without the usage of a dedicated GPU), it is a very cost-effective AR-based patient positioning method. It also opens the way for other researchers by introducing a framework that could be improved upon for better mobile interactive AR patient positioning solutions in the future.
\end{abstract}

\end{frontmatter}

\section{Introduction}

\begin{figure*}[t!]
    \centering
    
    \begin{tabular}{c c}
    \multirow[c]{2}{*}[5.8cm]{\subcaptionbox{\label{fig:qual_setup}}{\includegraphics[width=0.49\textwidth]{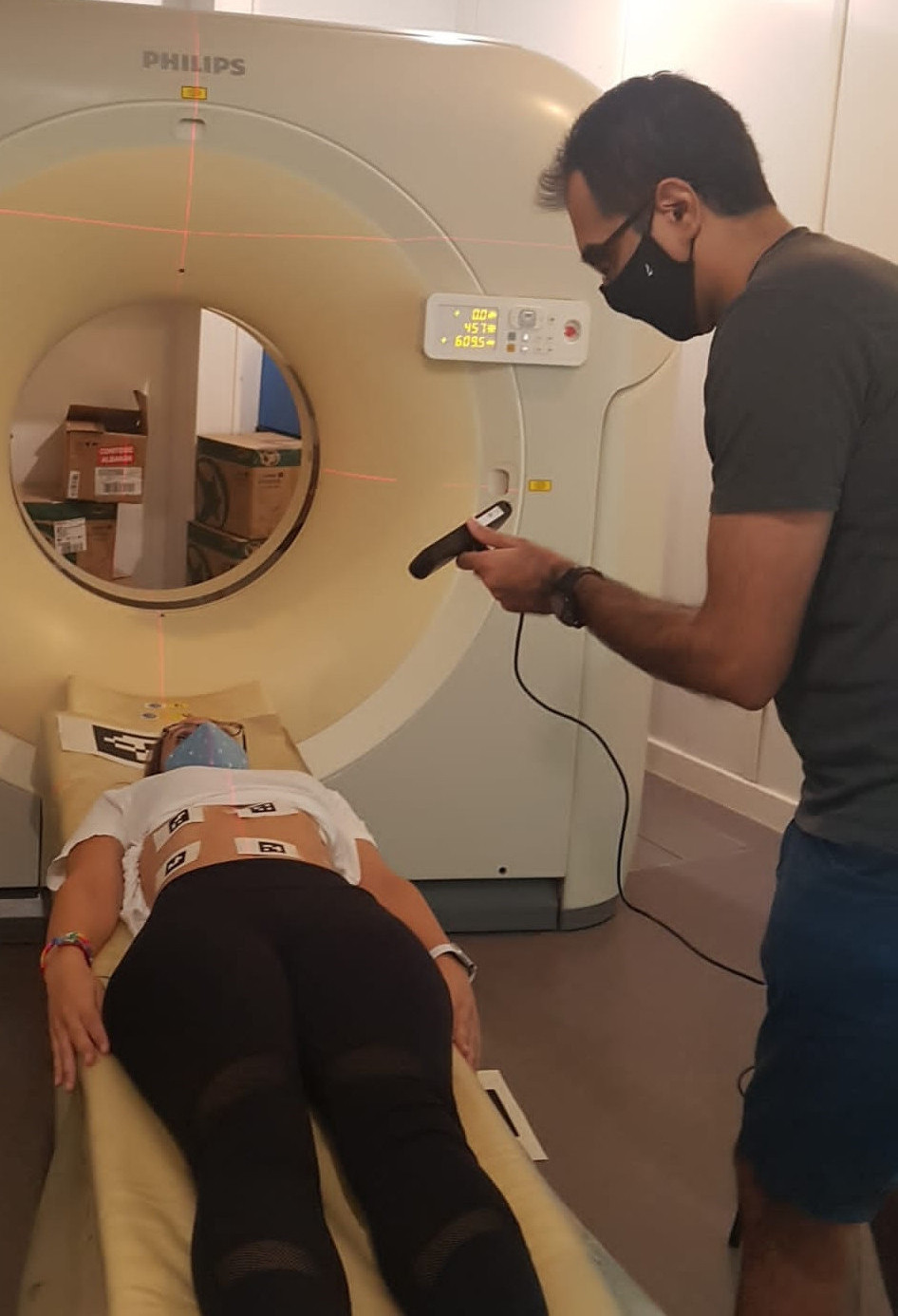}}}     & \subcaptionbox{\label{fig:raw}}{\includegraphics[width=0.49\textwidth]{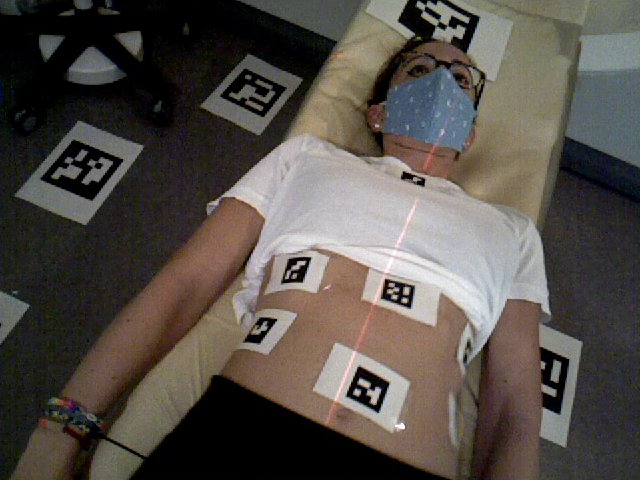}} \\
         & \subcaptionbox{\label{fig:AR}}{\includegraphics[width=0.49\textwidth]{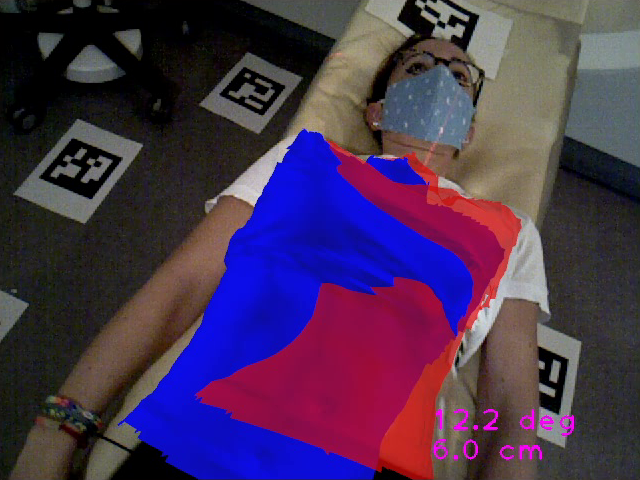}}
    \end{tabular}
    
    \caption{Setup employed for capturing data (\subref{fig:qual_setup}), and a frame from one of our testing sequences on a subject without (\subref{fig:raw}) and with (\subref{fig:AR}) augmented reality produced by our program. ArUco markers are attached to the body for tracking, while another set of ArUco markers are placed on the environment to  track  the camera pose in the environment. In \subref{fig:AR}, the red model mesh represents the tracked body of the patient, and the blue model mesh shows its desired pose. Our method also provides numerical feedback on the corner of the AR image in the form of rotational and translational error. }
    \label{fig:AR}
\end{figure*}

 The traditional process of radiation therapy is usually separated into two phases: the planning phase and the treatment phase. The treatment phase itself normally consists of multiple treatment sessions where the malignant tissue is radiated. In the planning phase, a CT scan is performed on the patient and, based on the result, the area to be radiated is planned for the subsequent radiation sessions in the treatment phase. It is therefore crucial that in the radiation session the position of the patient is the same as the position in the planning phase.

With the introduction of head-mounted displays such as the Microsoft Hololens\footnote{https://www.microsoft.com/en-us/hololens}, Augmented Reality (AR) has  gained traction in medical research,  specifically  in surgical applications such as training \cite{freschi_hybrid_2015,javaux_mixed-reality_2018} and intervention \cite{lee_calibration_2016,cartucho_multimodal_2020,luzon_value_2020}. Medical experts questioned about these systems have shown overwhelmingly positive opinions \cite{freschi_hybrid_2015,pelanis_use_2020}. However, the application of AR to guide patient positioning is so scarce that there is no related works in the recent literature reviews  \cite{eckert_augmented_2019,monsky_virtual_2019,tang_augmented_2020}. AR for patient positioning could have the potential benefit of assisting the operators by the interactive real-time visualization of the actual patient's position compared to the desired patient position (Fig.~\ref{fig:AR}).

In the past decade, there has also been a growth of works that take advantage of consumer-level depth or RGB-Depth (RGB-D) cameras (e.g. Microsoft Kinect) \cite{wang_poster_2015,macedo_high-quality_2014,ma_personalized_2016} which are very useful in AR applications. These sensors are affordable and also provide a real-time depth map of the scene and a corresponding color image. In other words they can give a dense real-time geometrical image of the scene rather than the sparse pose estimation that is possible using e.g. fiducial markers. Simultaneously, research on fiducial planar markers has proposed fast, robust and cheap methods for precise camera pose tracking. They do not need special equipment except for a color camera and a set of printed markers. These give fiducial planar marker detectors such as ArUco \cite{garrido-jurado_generation_2016,romero-ramirez_speeded_2018} many advantages with respect to the traditional infrared-based markers.

Taking advantage of these two recent technologies, this paper proposes a novel method for assisted patient positioning which is able to simultaneously track the patient and the treatment environment. Our system is able to render a virtual overlay of the patient's current pose and its desired pose using a freely moving RGB-D camera. Hence it can be employed for mobile interactive AR to guide patient positioning. To our knowledge this is the first method capable of such operation. Additionally, it is possible to take advantage of our approach for patient monitoring. This is possible by fixing the RGB-D camera pointing towards the patient. However this would not take advantage of the full potential of our method which lets the RGB-D camera move freely.

Our novel RGB-D based, model-based, object tracking algorithm is accurate and fast at the same time without the need of general purpose GPU computing on a dedicated GPU, using only the CPU unit. This makes our method usable on a wide range of hardware which makes it more accessible and cost-effective. We also believe this algorithm can have general applications beyond patient positioning or medicine. Our approach only requires an over the counter RGB-D camera and an average consumer laptop without the need for a powerful dedicated GPU. Although we have not seen any other similar work to our approach even in the industry, it is far more affordable than other industry level non-invasive surface-guided patient positioning methods such as AlignRT, Catalyst, and IDENTIFY \cite{hoisak_role_2018} that do not even have our AR capabilities.

The rest of this paper is structured as follows. Section \ref{sec:related_work} explains the  related works, then  in Section \ref{sec:methods} the proposed approach is introduced.  Section \ref{sec:experiment} describes the experimental results for validating our approach, after that those results are discussed in Section \ref{sec:discussion}, and finally Section \ref{sec:conclusion} draws some conclusions and future works.
\begin{table*}[t]
\centering

    \begin{tabular}{l||l|c|c|c|c|c}
        \multirow{2}{*}{Method} & \multirow{2}{*}{Sensor} & \multirow{2}{2cm}{Registration Speed (fps)} & \multirow{2}{*}{Precision (mm)} & \multirow{2}{1.5cm}{\centering Static Camera} & \multirow{2}{1.7cm}{\centering Regular Calibration} & \multirow{2}{1.7cm}{\centering Mobile AR} \\
        & & & & & & \\
        \hline\hline
        AlignRT & Proprietary & $\leq 5 $\cite{nguyen_commissioning_2020} & <1 \cite{wiencierz_clinical_2016}& Yes & Yes & No\\ \hline
        C-Rad Catalyst & Proprietary & Not Reported
        & 1  \cite{wiencierz_clinical_2016}& Yes & Yes & No\\ \hline
        C-Rad Sentinel & Proprietary & <1 \cite{stieler_clinical_2012} & 2  \cite{stieler_clinical_2012} & Yes & Yes & No
        \\ \hline
        Ehsani et. al. \cite{ehsani_registration_2019} & Kinect V2 & Not reported & 2 & Yes &  Yes & No
        \\ \hline
        Bauer et. al. \cite{bauer_multi-modal_2011} & Kinect V1 & Not reported & 12  & Yes & Yes & No
        \\ \hline
        Sarmadi et. al. \cite{sarmadi_3d_2019} & Xtion Pro Live & <1 & 11 &No & No & No\\ \hline
        Ours & Realsense L515 & 9 & 4 & No & No & Yes \\
    \end{tabular}
    
    \caption{Comparison of different patient positioning approaches.}
    \label{tab:comparison}
    
\end{table*}

\section{Related Work}
\label{sec:related_work}

\subsection{Patient Positioning in AR and Computer Vision}
From the few methods that investigate AR for patient positioning the early works do not perform any patient tracking and leave it to the user to detect when the pose is correct only with overlaying the desired pose of the patient on the video \cite{talbot_patient_2008,talbot_method_2009,french_augmented_2014}. These approaches need a calibration method to calibrate the fixed cameras with respect to the linear accelerator (linac). Another more recent method gives the possibility of using moving cameras however it still does not perform patient tracking and for that relies on the user's eyes \cite{cosentino_rad-ar_2017}. The most recent method we found that claims it has application in AR for patient positioning is \cite{ehsani_registration_2019}. They use a time of flight camera fixed to the linac and apply an advanced registration method to align the current patient's geometry to the reference model obtained in the planning phase. The mentioned methods either assume that the camera is fixed in the environment or they do not track the patient. A fixed camera limits the view in which the patient can be seen. Especially it makes the algorithm unsuitable to be used in conjunction with head-mounted displays (such as HoloLens) for AR. Another disadvantage is that the camera cannot show different parts of the patient on demand and it has to be fixed from before just for a specific point of view. Furthermore, the camera needs to be calibrated with respect to the linac and regularly checked if the calibration is still correct. Tracking the patient on the other hand, which is absent from most of the methods we mentioned, is necessary to give the user correct numerical or visual indications of the amount of the error in positioning the patient.

A new novel approach to computer-assisted patient positioning is presented in \cite{sarmadi_3d_2019}. They take advantage of a heightmap data structure reconstructed using their Global ICP algorithm and an RGB-D sensor. They use it to compare the pose of the patient in the planning phase and the treatment phase in radiation therapy. They also take advantage of ArUco markers \cite{garrido-jurado_generation_2016,romero-ramirez_speeded_2018} to align the reconstructed scenes. However, this approach is not capable of tracking the patient and giving visual feedback to the operator in real-time.

\subsection{Joint Scene and Object Tracking in Computer Vision}
Simultaneous tracking of the camera (scene) and the object in the scene is not a new idea. One early work is \cite{moosmann_joint_2013} where they apply self-localization and tracking of dynamic objects at the same time. The data is captured using a LIDAR in a self-driving car scenario. Here the scene is treated as a separate object. Another joint scene/object tracking algorithm has been employed for tracking people using data from a moving monocular camera \cite{choi_general_2013}.

One more recent work \cite{runz_maskfusion_2018} is a SLAM algorithm that recognizes moving object by segmenting them and treats the scene (background) just as another object. In this research, an RGB-D sensor is utilized for the detection and reconstruction of multiple rigid moving objects. Nevertheless, they need two dedicated GPUs one for performing the SLAM algorithm and another one just for the segmentation of objects using a convolutional neural network (CNN). Another similar approach is presented in \cite{strecke_em-fusion_2019}, however, in this approach they can additionally remove non-rigid moving objects from the background using a probabilistic framework for robust camera tracking. Notwithstanding they still need dedicated GPUs for object detection and reconstruction.

\subsection{Model-based Object Tracking In Computer Vision}
An early example of 3D model-based tracking is presented in \cite{comport_real-time_2006}. The authors present an algorithm that tracks a 3D object by tracking the contours of its projection on the image plane on using data from a monocular camera.

In \cite{choi_robust_2012} a particle-filter based tracking approach is presented that tracks the object using edge features. Keypoint features are employed for the initialization of tracking. They do not reach real-time performance and only suggest it could be possible with an implementation that takes advantage of GPU computing.

A Gaussian filter based tracking method using depth map input is put forward in \cite{issac_depth-based_2016}. Higher accuracy is achieved by robustification of the Gaussian filter and real-time performance is obtained by reducing the complexity of the filter. Despite that, no solution is provided for robust initialization or re-initialization of tracking which could be expected since they only use depth input.

A more recent approach is introduced in \cite{tsai_efficient_2018}. In this work CAD models of the object are employed in conjunction with reference pictures taken of them, however, their objects have simple shapes with planar surfaces. They use image feature matching to initialize a template matching algorithm and then deduce the 3D pose from that. Their algorithm is real-time however they require a discrete GPU.

In \cite{trinh_modular_2018} an approach for 3D tracking is presented that uses three different types of constraints: texture-based points, edges, and point-to-plane distance. Although their algorithm does not need any GPU acceleration it needs to use very simple geometries containing few surfaces for point-to-plane geometry-based registration which we think is not very appropriate for tracking complex shapes such as the human body.

\subsection{Comparison with other approaches}

This section compares, in Table \ref{tab:comparison}, our proposal  with  three of the most common commercial solutions (AlignRT, Catalyst and Sentinel) and other experimental solutions for patient positioning.

The table indicates, amongst other aspect, the type of sensor employed, the registration speed and precision. We would like to note that none of the methods are designed to be usable in mobile AR (except ours). In addition, our approach has a high registration speed, specially if we consider that no GPU is required.

Regarding the precision, AlignRT is the most precise method, while ours obtains a precision of $4$mm. It must be considered, though, that our method is designed to work with a moving camera, which is a more challenging problem since its position must be recalculated in every frame. Nevertheless, as we explain in the experimental section, there is room for improvement in the future as the depth camera technology evolves.

One advantage of our method  is that it does no need for regular recalibration like the others. There is only one other method that employs a moving camera (Sarmadi et. al. \cite{sarmadi_3d_2019}). However, they need to scan the person by moving the sensor to create only one frame of reconstruction which takes several second. Hence their method is not suitable for live feedback needed in AR.

Finally, we should mention an important disadvantage of the commercial methods which is their high price. As an example, the price of an AlignRT system starts from £150,000 plus £20,000 for annual service \cite{noauthor_nice_2018}. Other non-commercial proposed methods however normally use over the counter sensors, hence they have a low price. For example, the Intel Realsense L515 is priced at only \$349.

\begin{figure*}[t!]
    \centering
    \includegraphics[width=\textwidth]{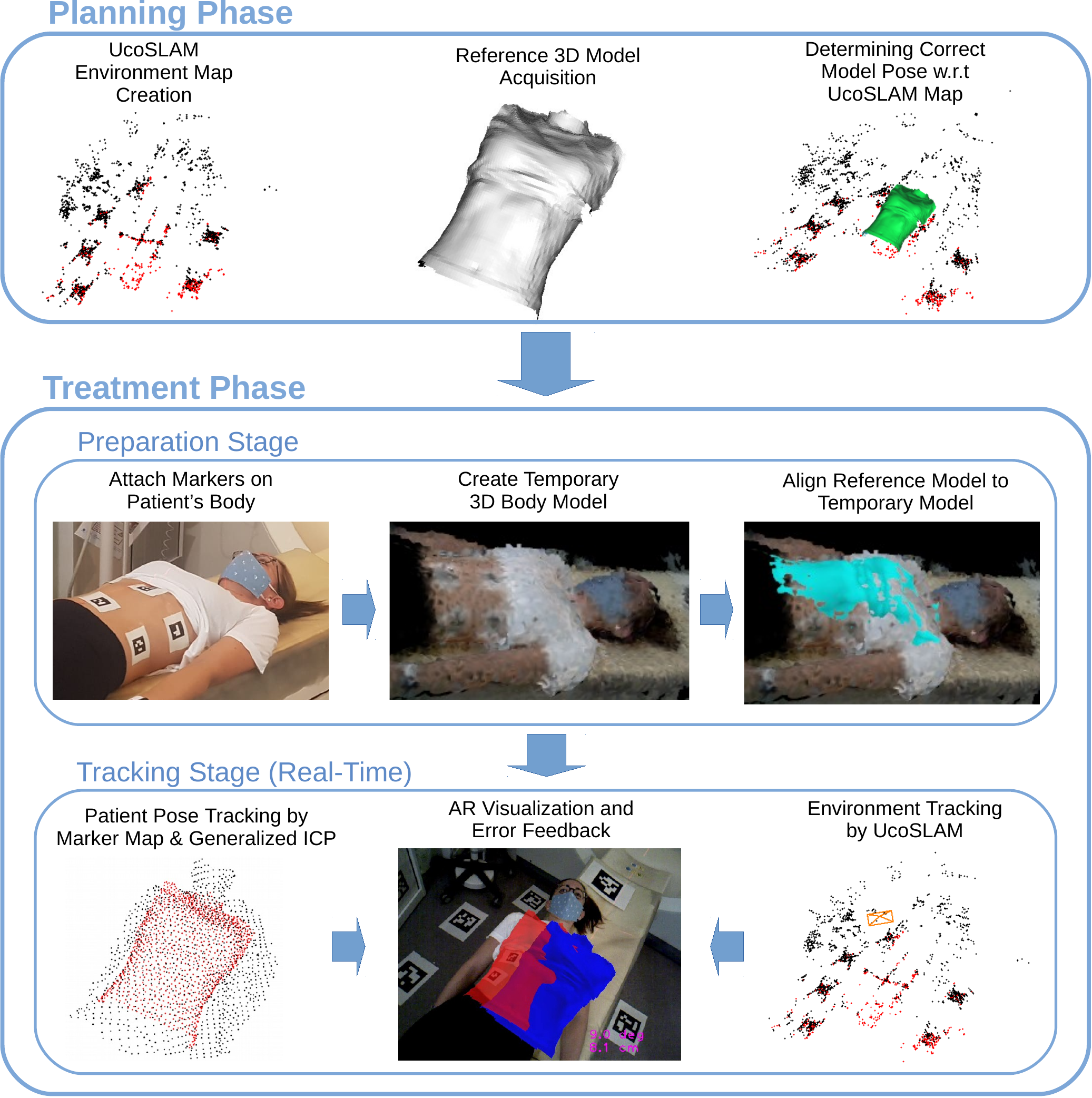}
    \caption{A high-level overview of our approach is presented. First, in the planning phase an UcoSLAM environment map of the treatment room, a 3D model of the body part, and the desired pose of the model with respect to the environment map are obtained. Then at each treatment session, in the preparation stage ArUco markers are attached to the patient's body, a temporary 3D body model is created and aligned to the reference model. In the tracking stage, which happens in real-time, the environment tracking by UcoSLAM is used to demonstrate the desired pose and patient pose tracking by marker map and generalized ICP is employed to show the patient's current pose. The desired pose is shown in blue color and the current pose is shown in red color in AR visualization. Finally numerical error feedback is added to the AR visualization.}
    \label{fig:overview}
\end{figure*}

\section{Methodology}
\label{sec:methods}
\subsection{Overview} 

Our proposed approach is designed to give real-time feedback for patient positioning in radiation therapy through AR to the person who performs the positioning. We assume that the non-rigid deformation of the patient's body matches that of the patient's desired pose. Then, our method enables viewing a virtual model of the patient overlaid on top of his/her body, and also a virtual model where the patient needs to move to. Furthermore, real-time quantitative feedback of the positioning error is shown to the operator in the augmented image. The mentioned information can be used for visualization in head-mounted displays or a tablet connected to the camera.

Figure \ref{fig:overview} provides a summary of our  approach which is explained in detail in this section.

Our method requires the following inputs obtained from the planning phase:
\begin{itemize}
 \item A map of the treatment room. This is obtained using the UcoSLAM \cite{munoz-salinas_ucoslam:_2019} algorithm, which employs 2D image features (keypoints) and ArUco planar markers \cite{garrido-jurado_generation_2016,romero-ramirez_speeded_2018} using an RGB-D camera. 
 \item A 3D reference model of the patient's body surface that can be obtained from a 3D laser scanner or  from the CT scan performed in the planning phase. 
 \item The correct pose of the 3D reference model w.r.t. the created environment map which indicates the desired pose of the patient in the treatment room. This could be done by a calibration method such as in \cite{talbot_method_2009}.
\end{itemize}
We assume that the steps corresponding to the planning phase have been carried out according to the mentioned methods.

In the treatment phase, for every session, the patient needs to wear tight-fitting clothes with small ArUco planar markers printed (similar to \cite{barbero-garcia_fully_2020}), or using stickers directly attached to the body with markers printed on them. We should mention that attaching the sticky markers is very easy and takes no more than a couple of minutes. One only needs to randomly place several of them in the body to make sure they are properly visible from the camera’s view point. While the markers do not need to be in the same position on the body from one treatment session to another, they must remain fixed within the session. In the first step of our algorithm, we create a temporary 3D body model of the patient including the 3D geometry and the positions of the markers on the patient's body. In the second step, we register the position of the markers (body marker map) to the accurate reference 3D model created in the planning phase. The third step is the only real-time step and happens while the operator is positioning the patient. In this stage, both the body of the patient and the environment are tracked with respect to the camera. The body is tracked by a combination of body marker map tracking and geometrical alignment of the reference 3D model on the depth map captured with the camera. The camera is tracked with respect to the environment with the UcoSLAM algorithm using the environment map created in the planning phase. For visualization, the tracked camera pose is employed to overlay the desired body pose (shown in blue in Figure \ref{fig:AR}\subref{fig:AR}) and the current body pose (shown in red). In addition, the difference between the desired body pose and the current patient's pose is calculated to show rotational and translational errors in patient position to the operator.

UcoSLAM is a new tracking algorithm that ourperforms other state-of-the-art SLAM algorithms in accuracy and speed as has been proved in \cite{munoz-salinas_ucoslam:_2019}. An advantage of this method is that it is the unique one that can map and track ArUco planar markers as well as image features (keypoints) as part of its design. Since markers can stay in the same place for long term, they are good reference points specially for our application where we want to track the environment consistently in different treatment sessions. Hence we have chosen the UcoSLAM algorithm to map and track the environment in the treatment room.

The  result of our system can be seen in Figure \ref{fig:AR}. The color image from the camera is shown with (Figure \ref{fig:AR}\subref{fig:AR}) and without (Figure \ref{fig:raw}) the augmented visualization. While the red transparent model represents the current patient's pose, the blue opaque model represents the desired one. The rotational and translation error of the positioning are also reported in the augmented image.

Since the main contribution of this paper is the process of simultaneously tracking the camera and the patient, we are not explaining in more details how to obtain the 3D reference model (e.g. from CT scan), create a UcoSLAM map of the treatment environment, and determine the target pose of the reference model with respect to the UcoSLAM map. These steps, which are done in the planning phase, are out of the scope of this paper and do not need any novel algorithm for their implementation.

The rest of this section provides a detailed explanation of the different steps involved in the proposed method. Creation of a temporary 3D body model of the patient is explained in Subsection \ref{sec:body_model}, alignment of the reference model to the temporary model is described in Subsection \ref{sec:ref_to_temp}, and finally the tracking stage is detailed in Subsection \ref{sec:tracking}.

\begin{figure*}[t!]
    \centering
    \includegraphics[width=\textwidth]{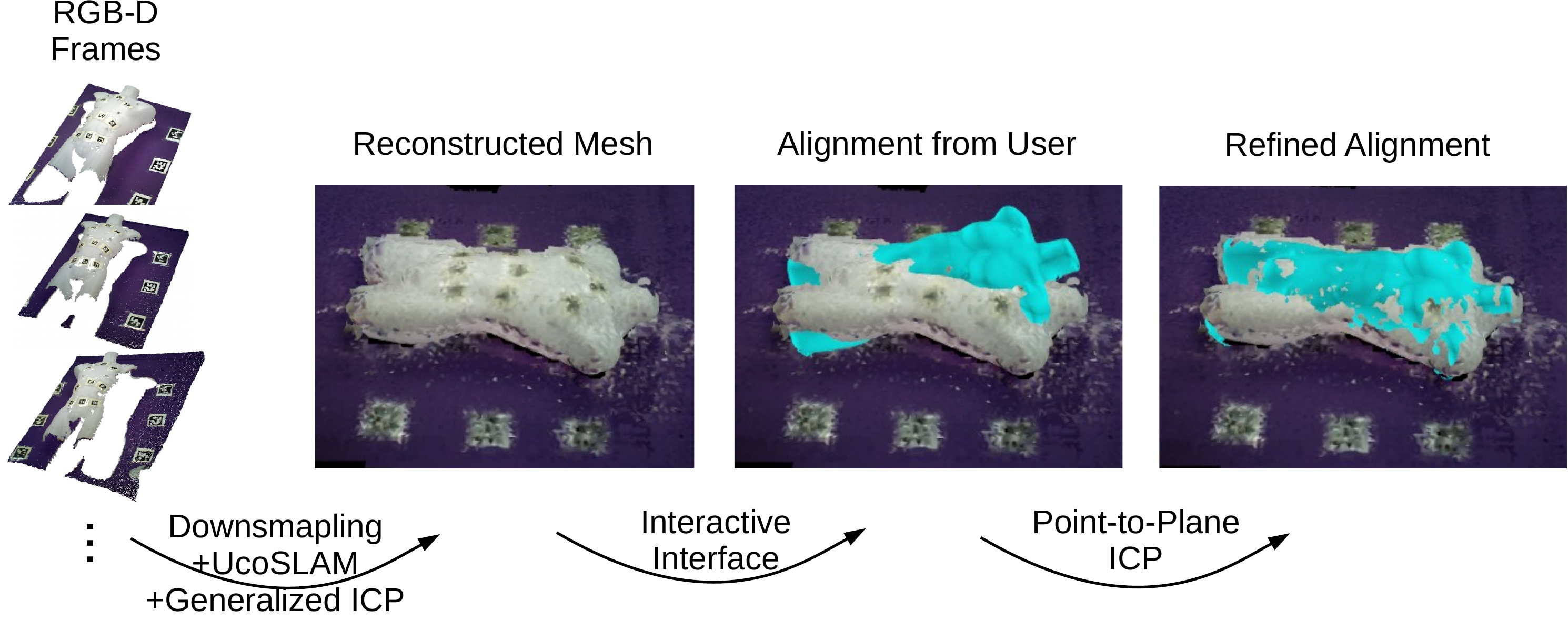}
    \caption{Visualization of temporary 3D reconstruction of the patient (represented by a mannequin) and its alignment to the given 3D model. First a rough alignment is given by the user and then it is refined furthermore by the point-to-plane ICP algorithm. Please note that only every l-th frame from the input RGB-D frame sequence is used for 3D reconstruction.}
    \label{fig:reconst_to_model}
\end{figure*}

\subsection{Temporary Body Model Creation}
\label{sec:body_model}

An RGB-D video sequence of the patient laying on the treatment bed is recorded to create a temporary 3D model of their body at the beginning of each treatment session. We refer to it as \textit{temporary} since it will only be used during the current treatment session. We propose a fast reconstruction method by combining  UcoSLAM \cite{munoz-salinas_ucoslam:_2019} and the generalized ICP \cite{segal_generalized-icp_2009} algorithm. Generalized ICP is a fast approximation of the point-to-plane ICP algorithm which we exploit to refine the result of UcoSLAM pose estimation. Generalized ICP is one of the fastest CPU based registration algorithms with a good accuracy compared to the state of the art \cite{parkison_semantic_2018}. The captured sequence is also used to create a three-dimensional map of the markers attached to the patient (\textit{body marker map}) \cite{munoz-salinas_mapping_2018} that is employed for tracking the patient with respect to the camera. 

Since the body marker map and temporary 3D model are obtained from the same video sequence, it is possible to establish the relationship between their reference system to align them. This is important since we will later need the alignment from the reference 3D model (CT scan) to the body marker map's coordinate system for correct pose estimation and visualization of the 3D model on top of the patient's body. 

Formally speaking, let us assume:

\begin{equation}
    F_i=(D_i,I_i), \quad i=1 \dots n
\end{equation}
represents the data in the $i$-th frame of the RGB-D sequence, where $D_i$ is the depth map and $I_i$ is the RGB image.

We feed the marker mapper algorithm \cite{munoz-salinas_mapping_2018} with the sequence of the images $\{I_i\}_{i=1}^{n}$ to create the marker map, $\mathcal{M}$. Since the marker mapper does not need time-coherent input, only a subset of the frames is employed to speed-up computation.

\begin{equation}
    \mathcal{M} = \textrm{MarkerMapper}(\{I_i|i\in \mathfrak{L}\})
\end{equation}
where
\begin{equation}
    \mathfrak{L} = \{i \in Z^+|1\le i\le n \;\land\; {i/\textrm{l}}=\lfloor{i/\textrm{l}}\rfloor\}
\end{equation}
Here, $\textrm{l} \in Z^+$ is a constant that enables processing only every l-th frame, and $\mathfrak{L}$ is the set of all valid indices.

The temporary geometrical reconstruction of the body and its alignment to the created body marker map are explained below.

\subsubsection{Geometrical Body Reconstruction}

Our geometrical body reconstruction algorithm takes advantage of both pose estimation from the UcoSLAM algorithm \cite{munoz-salinas_ucoslam:_2019} and the generalized ICP registration algorithm \cite{segal_generalized-icp_2009}. An example of this geometrical reconstruction can be seen in Figure \ref{fig:reconst_to_model}.

UcoSLAM is fed with the frame set $\{F_i\}_{i=1}^{n}$, generating as a result the set of 3D poses of the body w.r.t. the camera tracked for all frames, $\{P_i^U\}_{i=1}^{n}$ where $P_i^U$ is the $4\times 4$, 3D transformation matrix corresponding to the $i$-th.

We keep the body reconstruction in the form of a point cloud $\mathfrak{C}$ and  go through the frame indices $i\in \mathfrak{L}$  sequentially converting each depth map $D_i$ to a point cloud $C_i$:

\begin{equation}
    C_i=\textrm{DepthToPointcloud}(D_i)\;, \quad i\in\mathfrak{L}
\end{equation}
\begin{equation}
    \mathcal{C}_i=\textrm{Downsample}(C_i, d_i )\;,\quad i\in\mathfrak{L}
\end{equation}

Here the depth map $D_i$ is converted to a point cloud, $C_i$, using the known camera parameters and then downsmapled to $\mathcal{C}_i$ by voxel downsampling using a size of $d_i$, which is calculated dynamically for each frame as:

\begin{equation}
\label{eq:voxelsize}
    d_i=\max(\sqrt[3]{V_i/N},d_0)
\end{equation}
where $V_i$ is the volume of the bounding cube of $\mathcal{C}_i$, $N\in Z^+$ is a constant, and $d_0\in R^+$ is the minimum possible voxel size, also a constant.

To perform the reconstruction we transform each point cloud $\mathcal{C}_i$ to its corresponding reconstruction pose, $P_{i}^R$, and add it to $\mathfrak{C}$. 

In order to determine $P_{i}^R$ we take $P_i^U$ and refine it by the generalized ICP algorithm:

\begin{equation}
P_{i}^R= \left\{ \begin {array}{lcc} \textrm{GeneralizedICP}(\mathcal{C}_i,\mathfrak{C},P_i^U) &  \quad i\in \mathfrak{L} \land  i > 1 \\
\\ P_{1}^U &  i=1
\end{array}
\right.
\end{equation}

Again, not all frames are required for the reconstruction, thus, only a subset of them is employed to speed-up the computation. Also, since at the first frame $\mathfrak{C}$ is empty, there is no pose refinement done for the first frame and $P_{1}^R = P_1^U$.

To keep the size of the point cloud manageable, we also downsample $\mathfrak{C}$ after each addition:
\begin{equation}
    \mathfrak{C} \leftarrow \textrm{Downsample}(\mathfrak{C},d_{\mathfrak{C}})
\end{equation}
where $d_{\mathfrak{C}}\in R^+$ is a constant for the voxel size. 

\subsubsection{Alignment of Body Marker Map to Geometrical Body Reconstruction}

When processing the reconstruction sequence, we obtain the body marker map $\mathcal{M}$ but also its pose $P_i^M$ at each frame w.r.t. the camera:
\begin{equation}
    (P_i^M,S_i) = \textrm{MarkermapPose}(\mathcal{M}, I_i), \quad i\in\mathfrak{L}
\end{equation}
\noindent where $S_i \in \{true, false\}$ indicates whether marker map pose estimation was performed successfully.

Now we need to determine the transformation, $T$, that relates the coordinates systems of $\mathcal{M}$ and $\mathfrak{C}$. Let us define the set of all indices with successful marker map tracking by:
\begin{equation}
\mathfrak{S}=\{i\in\mathfrak{L}|  S_i = true\}
\end{equation}
Since we have corresponding poses for each frame where the marker map is successfully tracked we can write:

\begin{equation}
     P_i^R T = P_i^M, \quad i\in \mathfrak{S}
\end{equation}

Now it is possible to determine $T$ using the least square method:

\begin{equation}
     \left (\sum_{i\in \mathfrak{S}} {(P_i^R)}^\top P_i^R \right) T = \sum_{i\in \mathfrak{S}} {(P_i^R)}^\top P_i^M ,
\end{equation}
which is a system that can be solved for all columns of $T$ at the same time:

\begin{equation}
      T = \left (\sum_{i\in \mathfrak{S}} {(P_i^R)}^\top P_i^R \right)^{-1} \sum_{i\in \mathfrak{S}} {(P_i^R)}^\top P_i^M.
\end{equation}

To make sure that the rotation component in $T$ belongs to the SO(3) Lie group, we convert it to the axis angle representation and back:

\begin{equation}
\label{eq:T_matrix1}
    T=\begin{bmatrix}
     & R &  & \vec t \\
        t_{4,1} & t_{4,2} & t_{4,3} & t_{4,4} 
    \end{bmatrix}
\end{equation}

\begin{equation}
    (r,\theta) \leftarrow \textrm{AngleAxis}(R)
\end{equation}
\begin{equation}
    \hat R \leftarrow \textrm{RotationMatrix}(r,\theta)
\end{equation}
Here, $R$ is the $3\times3$ rotation component of $T$, $\vec t$ is its translation component, $r$ and $\theta$ are the rotation axis and angle obtained from $R$, and $\hat R$ is the rotation matrix created from $r$ and $\theta$.

We also fix the numbers on the last row of $T$ to create a transformation belonging to the SE(3) Lie group. Finally, we can write:
\begin{equation}
\label{eq:T_matrix2}
    \hat T=\begin{bmatrix}
     & \hat R &  & \vec t \\
        0 & 0 & 0 & 1 
    \end{bmatrix}
\end{equation}
where $\hat T$ is the final obtained transformation from the marker map coordinate system to the reconstruction coordinate system.

\subsection{Reference Model to Temporary Model Alignment}
\label{sec:ref_to_temp}

The alignment of the 3D reference model to the temporary body reconstruction is done semi-automatically. First, the user manually gives a rough alignment and then we use point-to-plane ICP to refine it. For visualization in the manual input we create a mesh out of our temporary body model's point cloud using Poisson surface reconstruction \cite{kazhdan_poisson_2006}:

\begin{equation}
    \mathfrak{M}_\textrm{rec} \leftarrow \textrm{PoissonSurface}(\mathfrak{C})
\end{equation}
where $\mathfrak{M}_\textrm{rec}$ is the triangle mesh surface created from the temporary model point cloud, $\mathfrak{C}$. It should be mentioned that we prune the resulting mesh so that it does not have vertices that are too far away from any point in $\mathfrak{C}$.

We denote the manual alignment given by the user by $T_\textrm{manual}$ and refine this transformation by:
\begin{equation}
    T_\textrm{refined}=\textrm{PointToPlaneICP}( \mathcal{V}_\textrm{rec},\mathcal{V}_\textrm{model},T_\textrm{manual})
\end{equation}

Here $\mathcal{V}_\textrm{model}$ is the set of vertices from the patient's reference model mesh, $\mathcal{V}_\textrm{rec}$ is the set of vertices from the patient's temporary reconstruction mesh, and $T_\textrm{refined}$ is the result of the point-to-plane ICP alignment of $\mathcal{V}_\textrm{rec}$ to $\mathcal{V}_\textrm{model}$, initialized by $T_\textrm{manual}$. A visual example of 3D reconstruction to model alignment from our implementation can be seen Figure \ref{fig:reconst_to_model} in the first two images from right.

Finally to obtain the alignment from the marker map $\mathcal{M}$ to the reference model we can write:

\begin{equation}
    \mathfrak{T}=T_\textrm{refined}\hat{T}
\end{equation}
\noindent where $\hat{T}$ was defined in Eq.~\ref{eq:T_matrix2}, and $\mathfrak{T}$ is a transformation that takes a point from the body marker map coordinate system to the reference model coordinates system. Now it is possible to perform our hybrid model based patient tracking taking advantage of $\mathfrak{T}$.

\begin{figure*}[t]
    \centering
    \includegraphics[width=\textwidth]{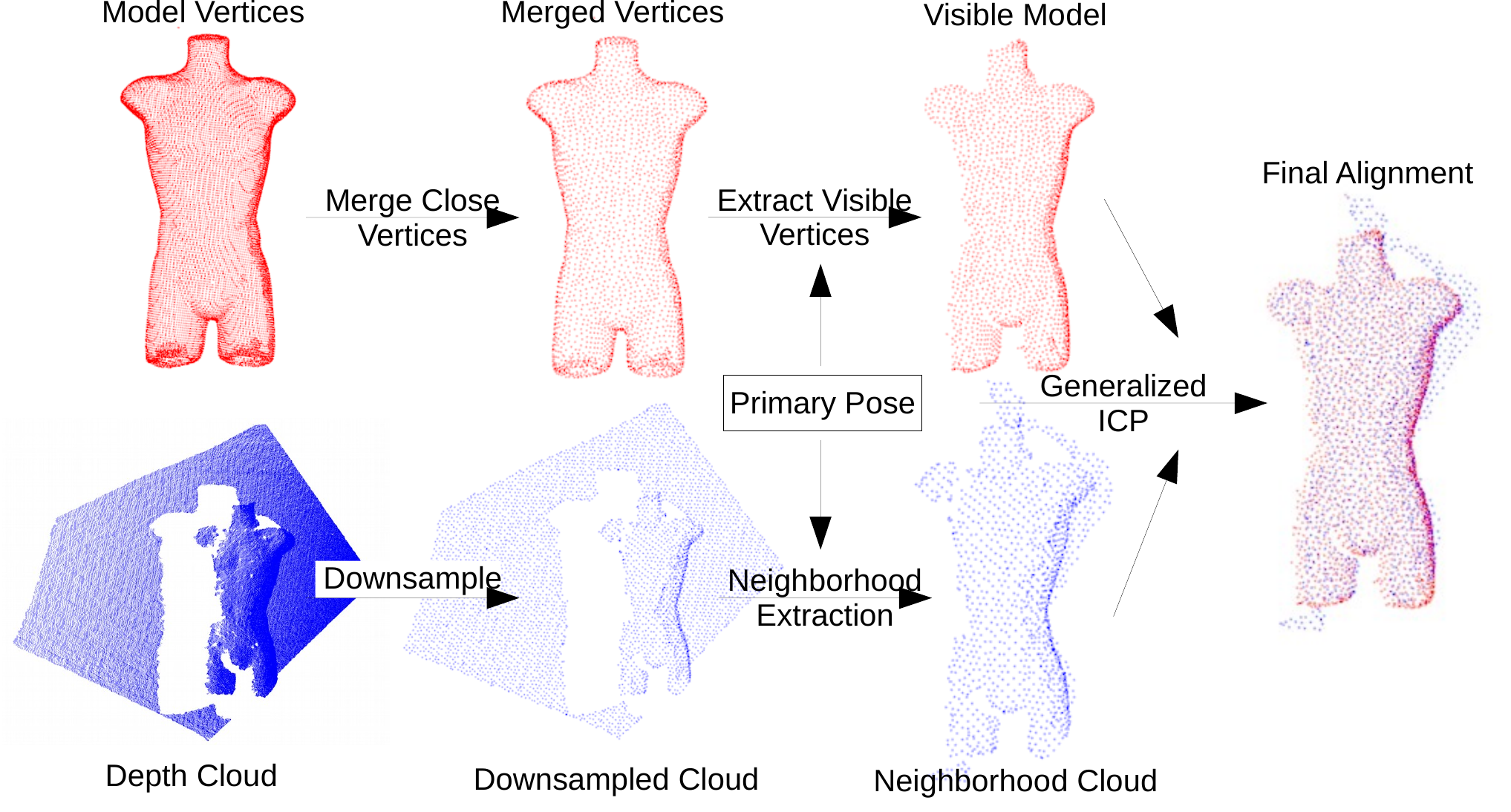}
    \caption{Steps of geometric alignment in our tracking algorithm. Please note that merging model vertices happens only one time before tracking starts. The rest of the steps need to be repeated for individual frames.}
    \label{fig:geometric_alignment}
\end{figure*}

\subsection{Tracking}
\label{sec:tracking}

Our tracking algorithm consists of two main parts: scene tracking and patient tracking. Scene tracking is done by the UcoSLAM algorithm in tracking mode. This approach needs a reconstruction of the scene in the form of a UcoSLAM map with the patient not being present. This reconstruction should be done in the planning phase of radiation therapy so that it is consistent between treatment sessions. The reason the patient should not be present while doing scene reconstruction is that it could confuse scene tracking later when the patient is present and moving in the scene.

Let us assume:
\begin{equation}
 F^*_i=(D^*_i,I^*_i), \quad i=1,\dots,m   
\end{equation}
represents the data in the i-th frame of the tracking sequence where $D^*_i$ is the depth map and $I^*_i$ is the RGB image. At each frame, first, we track the pose of the scene using the UcoSLAM algorithm:

\begin{equation}
P^{U*}_i \leftarrow \textrm{UcoSLAM}(F^*_i)
\end{equation}
where $P^{U*}_i$ is the scene pose in real-time at frame $i$ which is used to visualize the patient in its desired pose. This is useful for error calculation and visual feedback while performing the positioning. We also need to calculate the pose of the patient for proper visual feedback to the user.

\subsubsection{Patient Tracking}

Pose estimation of the patient is first done by ArUco marker map tracking \cite{garrido-jurado_generation_2016,romero-ramirez_speeded_2018} of the created body marker map \cite{munoz-salinas_mapping_2018}. Then the pose is further refined employing the generalized ICP \cite{segal_generalized-icp_2009} algorithm, registering the 3D geometry of the reference model to the depth map in the current frame. Before this registration, some preparations are done on the 3D data to improve the result which are discussed below and are also visualized in Figure \ref{fig:geometric_alignment}.  

\begin{table*}[t!]
\begin{threeparttable}
    \centering
    \begin{tabular}{|c|c|l|}
        \hline
        Parameter &  Value & Description\\ \hline \hline
        l & 10 & Constant for skipping frames in marker mapper and temporary model reconstruction. \\ \hline
        $N$ & \num{5e4} & Constant for dynamic subsampling\\ \hline
        $d_0$ & 1 cm & Minimum voxel size for dynamic subsampling \\ \hline
        $d_{\mathfrak{C}}$ & 1 cm & Downsampling voxel size for body reconstruction\\ \hline
        $d_*$ & 2 cm\tnote{1} , 1 cm\tnote{2} & Downsampling voxel size for depth point cloud in patient tracking\\ \hline
        $d_\textrm{nei}$ & 10 cm\tnote{1} , 4 cm\tnote{2} & Neighborhood extraction maximum distance in patient tracking \\ \hline
    \end{tabular}
    \begin{tablenotes}
    \footnotesize
    \item[1] Used with the Xtion Pro Live dataset
    \item[2] Used with the Realsense L515 dataset
    \end{tablenotes}
    \caption{List of values we used for different constants in our algorithm for quantitative evaluation.}
    \label{tab:parameters}
\end{threeparttable}
\end{table*}

We denote the body marker map pose estimation by:

\begin{equation}
    (P_i^{M*},S_i^*) = \textrm{MarkerMapPose}(\mathcal{M}, I^*_i)
\end{equation}
Here, again $S_i^* \in \{true,false\}$ determines if the pose estimation is performed successfully and $P_i^{M*}$ is the estimated pose. 

Marker map pose estimation is prone to errors that can happen due to motion blur or imperfect marker map to model alignment. We compensate this by refining the pose estimation using the data from the depth map.

First of all, we need to create a point cloud from the depth map and then downsample it, similar to temporary body reconstruction:

\begin{equation}
    C^*_i=\textrm{DepthToPointcloud}(D^*_i) 
\end{equation}
\begin{equation}
    \mathcal{C}^*_i=\textrm{Downsample}(C^*_i, d_* )
\end{equation}
where $C^*_i$ is the point cloud created from $D^*_i$, $\mathcal{C}^*_i$ is the downsampled point cloud and $d_*\in R^+$ is the voxel size for voxel downsmapling which is a constant.

Similar to temporary model reconstruction, we define the set of frame indices where the marker map performs pose estimation successfully:

\begin{equation}
\mathfrak{S^*}=\{1 \leq i \leq m |  S^*_i = true\}
\end{equation}

Let us assume that $\textrm{k}$ is the first frame number where the marker map pose estimation is successfully performed:
\begin{equation}
\textrm{k} = \min \{i \in \mathfrak{S^*}\}
\end{equation}
Then we define the primary pose of the patient in the current frame:
\begin{equation}
P_i^*=\left\{ 
\begin{array}{l l} 
(P_i^{M*}) \mathfrak{T}^{-1} & i\in\mathfrak{S^*} \\
P_{i-1}^* & i\notin\mathfrak{S^*}\land i>\textrm{k}
\end{array} \right.
\end{equation}

To increase the speed of our algorithm, we reduce the number of vertices in our input model mesh:
\begin{equation}
\mathfrak{M}^*_\textrm{model} = \textrm{MergeCloseVertices}(\mathfrak{M}_\textrm{model},d_*)
\end{equation}
where we merge neighboring vertices in model mesh, $\mathfrak{M}_\textrm{model}$, that are closer than the constant $d_*$. The merged vertices are replaced with a single vertex with the average of their position. The advantage of merging vertices instead of subsampling them by voxels is that the surface structure can be better preserved since merging is applied to neighboring vertices on the mesh. We define the vertices in $\mathfrak{M}^*_\textrm{model}$, as $\mathcal{V}_\textrm{model}^*$ and use them to refine the alignment of the pointcloud from depthmap, $\mathcal{C}^*_i$.

In order to increase the speed and accuracy of pose refinement, we remove the points in $\mathcal{V}_\textrm{model}^*$ which are not visible according to the primary pose $P_i^*$ with respect to the camera. We use the algorithm in \cite{katz_direct_2007} for this purpose:
\begin{equation}
\mathcal{V}_i^\textrm{visible}=\textrm{ExtractVsibilePoints}(\mathcal{V}_\textrm{model}^*,t_i^\textrm{cam})
\end{equation}
where $t_i^\textrm{cam}$ is the relative position of the camera with respect to the model:
\begin{equation}
     \begin{bmatrix}
     & R_i^\textrm{cam} &  & t_i^\textrm{cam} \\
        0 & 0 & 0 & 1
    \end{bmatrix}_{4\times4}=T_i^*
\end{equation}
\begin{equation}
    T_i^*=(P_i^*)^{-1}
\end{equation}

To increase the registration speed we also remove the parts of the pointcloud $\mathcal{C}^*_i$ that are not close to the model according to the primary pose:

\begin{equation}
\begin{multlined}
    \mathcal{C}^\textrm{nei}_i=\\ \left \{p\in \mathcal{C}^*_i \left | \exists p' \in \mathcal{V}_\textrm{model}^* : \left \|T_i^*\begin{bmatrix} 
p\\1\end{bmatrix}-\begin{bmatrix} 
p'\\1\end{bmatrix}\right \|_2<d_\textrm{nei} \right . \right \}
\end{multlined}
\end{equation}
Here $\mathcal{C}^\textrm{nei}_i$ is the cloud that contains the neighborhood of the patient's body in the primary pose in $\mathcal{C}^*_i$ and $d_\textrm{nei} \in R^+$ is the distance we use to determine it. 

Finally, we can refine the pose of the patient using the point cloud $\mathcal{C}^\textrm{nei}_i$:

\begin{equation}
    \hat T_i^* = \textrm{GeneralizedICP}(\mathcal{C}^\textrm{nei}_i,\mathcal{V}_i^\textrm{visible},T_i^*)
\end{equation}
Now we can assign the final pose to be used for visualization by:
\begin{equation}
 \hat P_i^* = (\hat T_i^*)^{-1}
\end{equation}

At the end to calculate the final pose error we can write:
\begin{equation}
 P_i^\textrm{ adj} = \hat {P_i^*}\left (P_i^{U*}\;P_U^\textrm{ref}\right )^{-1}
\end{equation}
where $P_U^\textrm{ref}$ is the reference pose of the patient with respect to the UcoSLAM map that we assume is given by the user through an interface, and $P_i^\textrm{ adj}$ is the transformation that can be used to give feedback to the person about the error in positioning. In the scenario of patient positioning $P_U^\textrm{ref}$ is determined in the planning phase of radiation therapy.

\section{Experimental Results}
\label{sec:experiment}
This section aims at evaluating the validity of the proposed method, both qualitatively and quantitatively, for patient positioning. 

The quantitative evaluation has been done using a mannequin as the patient. The tracking accuracy of the proposed method was measured with a motion capture system along with the running speed of our implementation. In the qualitative evaluation,  human subjects are employed and the output of our algorithm is demonstrated for visual inspection.

This section presents first the implementation details of our algorithm, including the hardware and software libraries employed, as well as the values of the employed parameters. Then, we demonstrate numerical results related to our quantitative evaluation. Finally, the qualitative results are presented using snapshots of our program's video output.
\subsection{Implementation Details}

We tested our implementation on a laptop with Intel® Core™ i7-4700HQ  running the Ubuntu 18.4 operating system. To capture RGB-D images we used the Asus Xtion Pro Live and the Intel Realsense L515 sensors. Both the depth and RGB images of the Xtion sensor were set to the resolution of $640\times480$ pixels. The depth image of L515 was set to the $640\times480$ resolution however the color had the resolution of $1280\times720$. The depth cameras were manually calibrated with respect to the color cameras and depth images were registered to the color camera coordinate system.

Our algorithm was implemented in C++ and the 3D visualization was developed using the Qt3D\footnote{https://wiki.qt.io/Qt3D} V5.9 library. For general image processing, we took advantage of OpenCV\footnote{https://opencv.org/} V3. To perform point cloud and mesh processing, including the point-to-plane ICP we employed the Open3D\footnote{http://www.open3d.org/} V0.9.0 library. We also took advantage of the original implementation of Generalized ICP which is publicly available online\footnote{https://github.com/avsegal/gicp}.

We also employed publicly available implementations of UcoSLAM\footnote{https://sourceforge.net/projects/ucoslam/} V1.0.8, marker mapper\footnote{https://sourceforge.net/projects/markermapper/} V1.0.15, and ArUCO\footnote{https://sourceforge.net/projects/aruco/} V3.1.11. Finally, the KinectFusion algorithm \cite{newcombe_kinectfusion:_2011} has been employed to create the 3D model reference of the mannequin employed in our tests. 

Along with the paper, we have employed several parameters for our algorithm. Their concrete values employed in our quantitative experimentation are indicated in Table \ref{tab:parameters}.

\subsection{Quantitative Evaluation}

\begin{figure*}[t!]
    \centering
    \begin{tabular}{c c}
        \multirow{2}{*}[3cm]{\subcaptionbox{\label{fig:setup}}{\includegraphics[width=.5\textwidth]{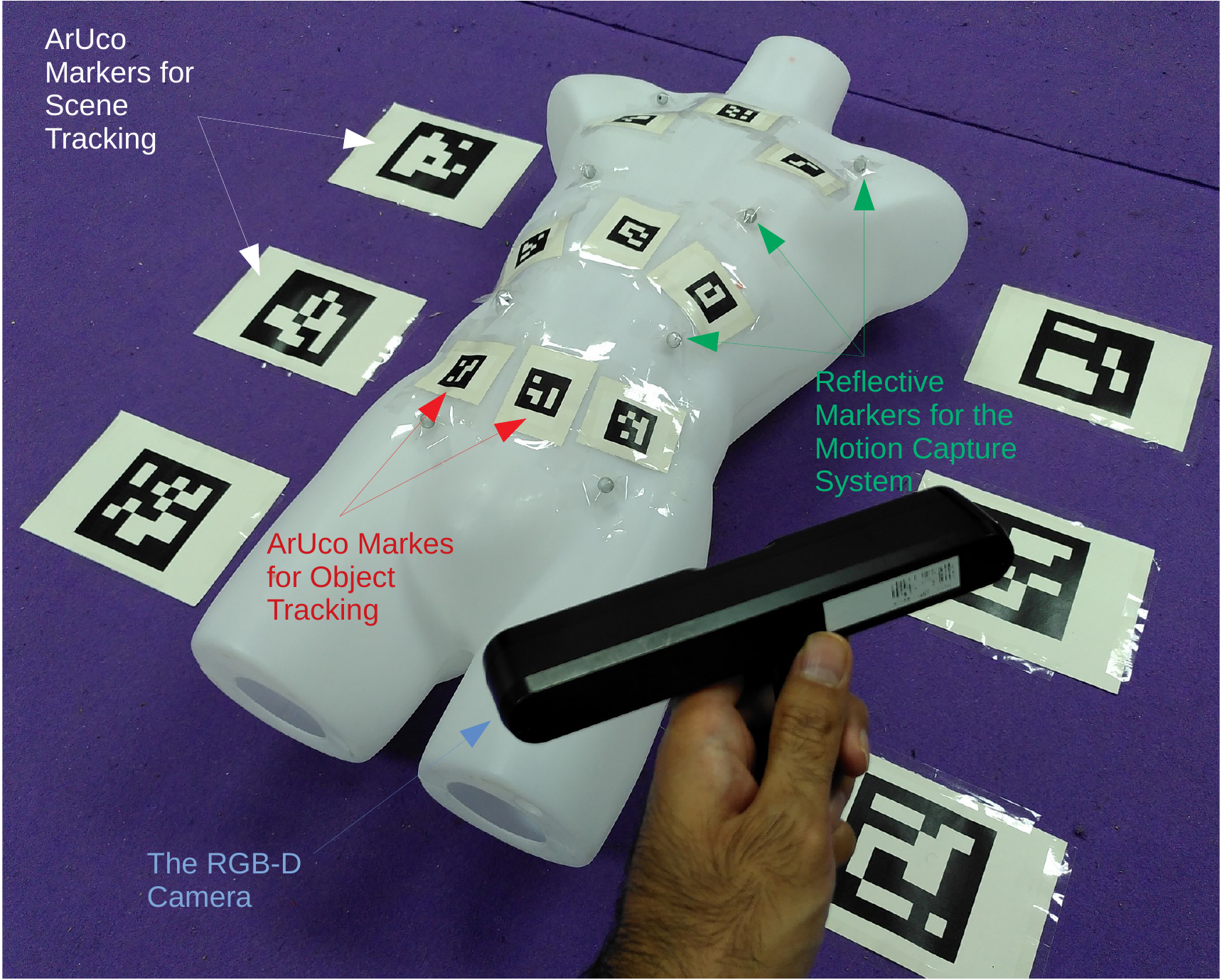}}} & \subcaptionbox{\label{fig:l515}}{\includegraphics[width=.5\textwidth]{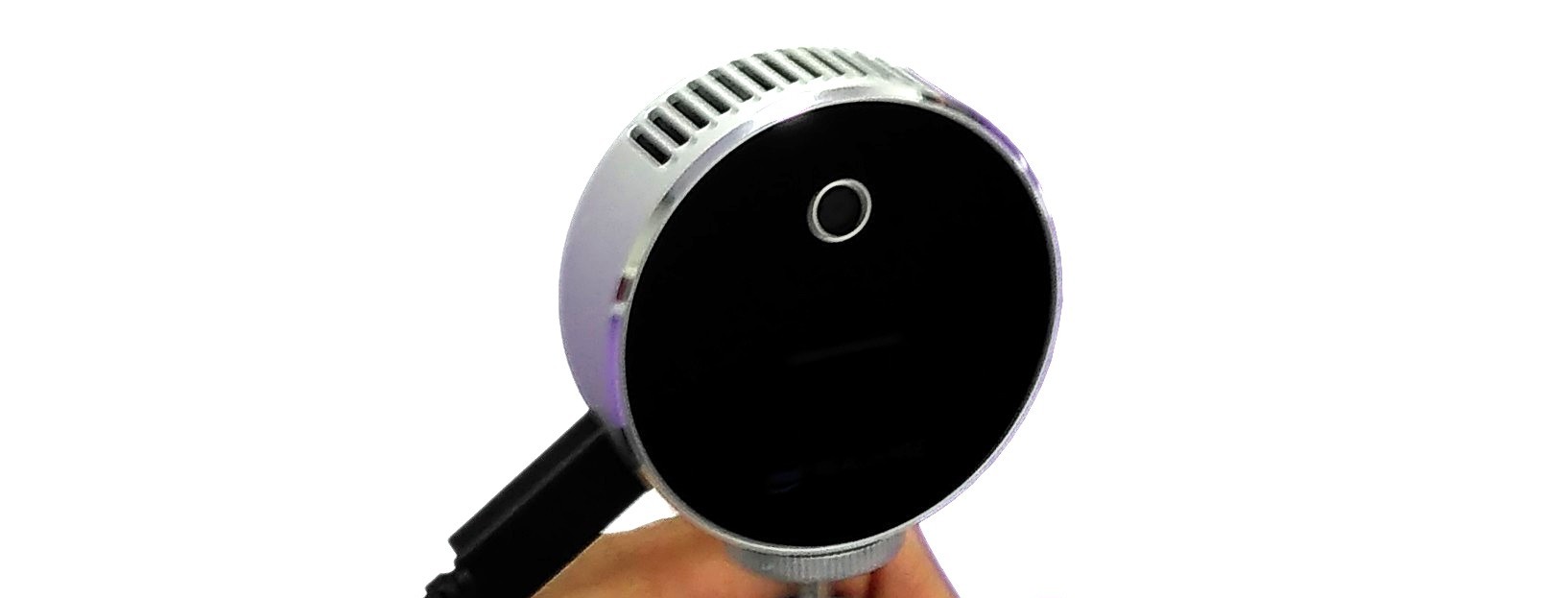}}
        \\
         & \subcaptionbox{\label{fig:xtion}}{\includegraphics[width=.5\textwidth]{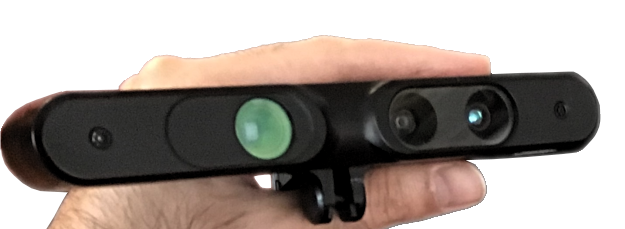}}
         \\
    \end{tabular}
    
    \caption{A demonstration of the setup for our quantitative experiment is presented (\subref{fig:setup}). Infrared reflective markers were tracked by the motion capture system for estimating the ground truth pose. The RGB-D camera was employed to capture the test video sequence. We took advantage of the Realsense L515 (\subref{fig:l515}) and the Xtion Pro Live (\subref{fig:xtion}) RGB-D cameras in our evaluation.}
    \label{fig:experiment_setup}
\end{figure*}

The quantitative evaluation has been carried out using a mannequin (see Fig. \ref{fig:experiment_setup}(\subref{fig:setup})) with infrared reflective dots attached. The dots are tracked using an OptiTrack\footnote{https://www.optitrack.com/} motion capture system comprised by a total of six infrared synchronized cameras that achieves sub-millimeter precision in the estimation of the dot positions. The reflective markers are used to determine the ground truth poses (rotation and translation) of the mannequin along the test sequences recorded. We also placed ArUco markers on the floor for accurate scene tracking using UcoSLAM, which combines ArUco markers, image features, and depth to track the camera pose in the environment. 

We captured two datasets, one with the Xtion Pro Live sensor and another one using the Realsense L515 sensor. For each dataset, we recorded a first RGB-D video sequence of the mannequin to obtain the reference 3D model using the KinectFusion algorithm, and another sequence of the environment without the mannequin to create the UcoSLAM environment map. These sequences are the equivalent of our planning phase.

Then, 6 sequences were recorded for evaluation purposes, where the mannequin is moved and rotated in different poses around the target position. Each sequence lasts several seconds and the total amount of frames from all sequences is $3006$ for each type of sensor that was employed. To simulate a real scenario and to properly analyze the system accuracy, the sequences were recorded with the camera and the mannequin in different relative poses and moving them locally during the video sequences. We tried to keep enough amount of the background (scene) in the captured images to make sure the UcoSLAM scene tracking is performed correctly.  

In order to estimate the relationship between the motion capture system reference system and ours, we split each tracking sequence into two halves. The first half was used to estimate the essential transformations needed for evaluation and the second half was employed for calculating the errors with respect to the ground truth.

We computed the error both in translation and rotation, and the results can be seen in Table \ref{tab:quantitative} for the datasets related to each type of sensor. We evaluated the mean error for each of our 6 tracking sequences. We also calculated these error values on the collection of all evaluation frames from all sequences ($1503$ frames for each sensor), the result of which can be observed in the last row of the table. 
\begin{table}[]
    \centering
    \begin{tabular}{c||c|c||c|c}
         \multirow{3}{*}{Sequence \#} & 
         \multicolumn{2}{c||}{\multirow{2}{2.3cm}{Asus Xtion Pro Live}}
          & \multicolumn{2}{c}{\multirow{2}{2.3cm}{Intel Realsense L515}} \\ 
          & \multicolumn{2}{c||}{} & \multicolumn{2}{c}{}
          \\ \cline{2-5}
          & MRE & MTE & MRE & MTE
          \\\hline \hline
         1 & \SI{1.77}{\degree} & \SI{6.87}{\milli\meter} & \SI{0.82}{\degree} & \SI{4.59}{\milli\meter}\\ \hline
         2 & \SI{1.45}{\degree} & \SI{5.27}{\milli\meter} & \SI{0.78}{\degree} & \SI{3.45}{\milli\meter}\\ \hline
         3 & \SI{2.19}{\degree} & \SI{8.67}{\milli\meter} & \SI{1.05}{\degree} & \SI{4.89}{\milli\meter}\\ \hline
         4 & \SI{1.81}{\degree} & \SI{7.51}{\milli\meter} & \SI{0.55}{\degree} & \SI{4.02}{\milli\meter}\\ \hline
         5 & \SI{1.65}{\degree} & \SI{6.94}{\milli\meter} & \SI{0.87}{\degree} & \SI{3.23}{\milli\meter}\\ \hline
         6 & \SI{1.73}{\degree} & \SI{8.41}{\milli\meter} & \SI{0.86}{\degree} & \SI{4.64}{\milli\meter}\\\hline\hline
        \multirow{2}{2.3cm}{\centering All Evaluation Frames} & \multirow{2}{*}{\SI{1.77}{\degree}} & \multirow{2}{*}{\SI{7.28}{\milli\meter}} & \multirow{2}{*}{\SI{0.82}{\degree}} & \multirow{2}{*}{\SI{4.17}{\milli\meter}}\\
        & & & &
    \end{tabular}
    \caption{Mean rotational error (MRE) and mean translational error (MTE) of our algorithm applied on the datasets captured by the Asus Xtion Pro Live and Intel Realsense L515 sensors. The evaluation is done by comparing the output to the ground truth from the motion capture system.}
    \label{tab:quantitative}
\end{table}
Furthermore, we calculated the overall median errors for both of the datasets as \SI{6.71}{\milli\meter}/\SI{1.53}{\degree} for the Xtion Pro Live sensor and \SI{3.83}{\milli\meter}/\SI{0.77}{\degree} for the Realsense L515 sensor.

We have also evaluated the running speed of our implementation. We calculated the average running time in the form of frames-per-second (fps) for all frames of all tracking sequences used for evaluation. To do so, for each frame, we measured the time lapsed since the previous pose estimation until the current pose estimation. The mean running speed of the patient tracking was 19 fps for the Xtion sensor and 9 fps for the Realsense L515 sensor

Before being able to run the program for patient tracking it is needed to create a marker map of the patient and align it to the 3D model. This requires a few steps that we call the preparation steps. We have summarized the running time for those steps that take a significant time, in Table \ref{tab:preparation_time} using our reconstruction sequences.

\begin{table*}[t]
    \centering
    \begin{tabular}{c||c|c}
       \multirow{2}{*}{Preparation Step} & \multirow{2}{2cm}{\centering Asus Xtion Pro Live} & \multirow{2}{2.3cm}{\centering Intel Realsense L515} \\
       & & \\ 
       \hline \hline
        Map Creation: UcoSLAM + Marker Map & \SI{37}{\second} & \SI{62}{\second}\\ \hline
        Geometrical Reconstruction & \SI{37}{\second} & \SI{68}{\second}\\ \hline
        Mesh Creation & \SI{2}{\second} & \SI{2}{\second}\\ \hline
        Interactive Reference Model Alignment& \SI{10}{\second} & \SI{31}{\second}\\ \hline \hline
        Total & \SI{86}{\second} & \SI{163}{\second}
    \end{tabular}
    \caption{The time spent by each part of the algorithm to create the marker map and UcoSLAM map in addition to finding the transformation form the marker map to the 3D model coordinate system.}
    \label{tab:preparation_time}
\end{table*}

\subsection{Qualitative Evaluation}

\begin{figure*}[p!]
    \centering
    
    \includegraphics[width=.3\textwidth]{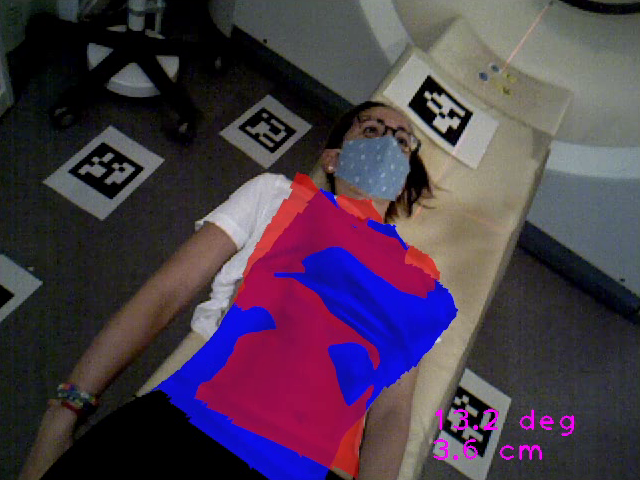}
    \includegraphics[width=.3\textwidth]{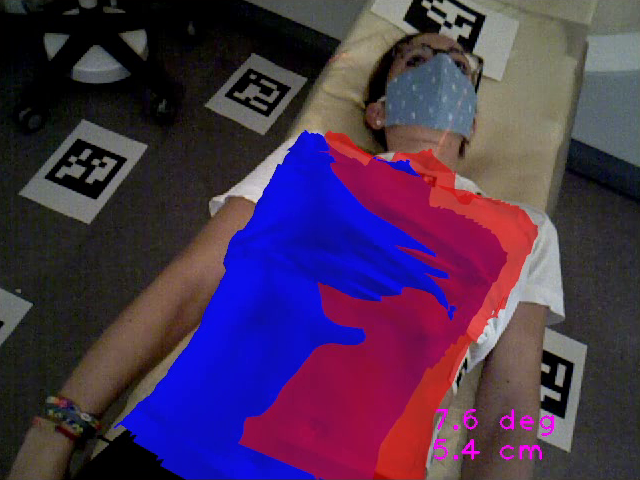}
    \includegraphics[width=.3\textwidth]{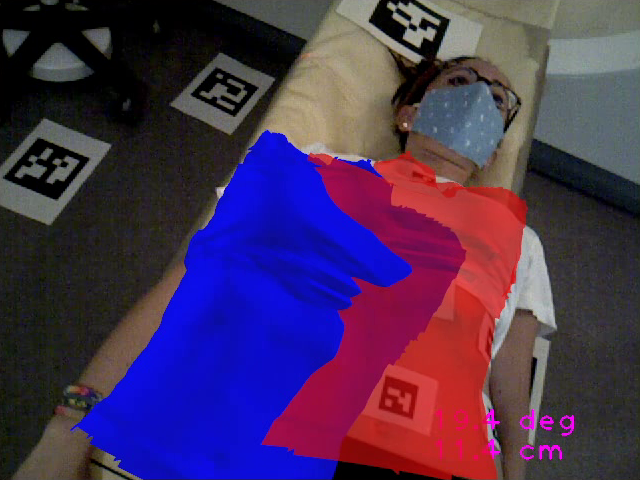}\\
    \includegraphics[width=.3\textwidth]{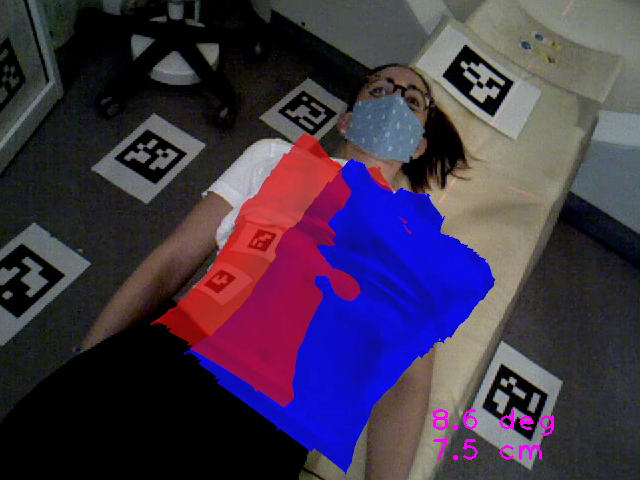}
    \includegraphics[width=.3\textwidth]{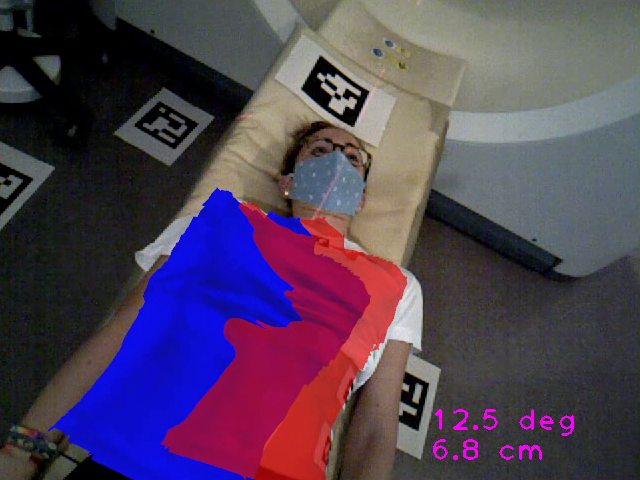}
    \includegraphics[width=.3\textwidth]{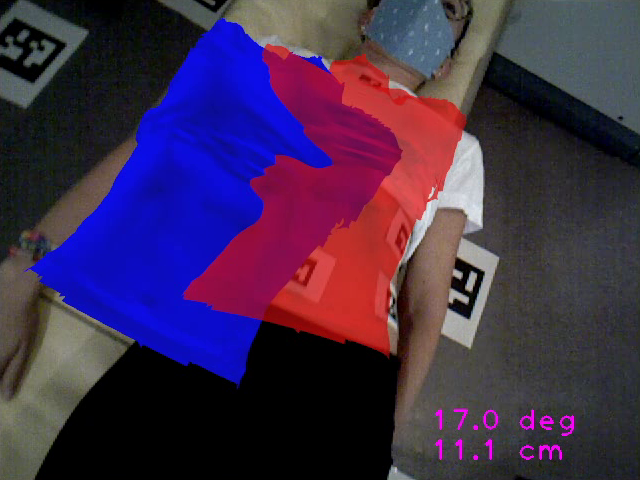}\\
    \includegraphics[width=.3\textwidth]{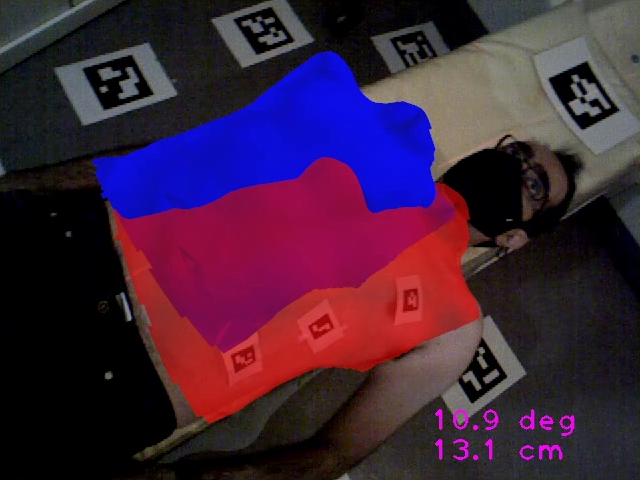}
    \includegraphics[width=.3\textwidth]{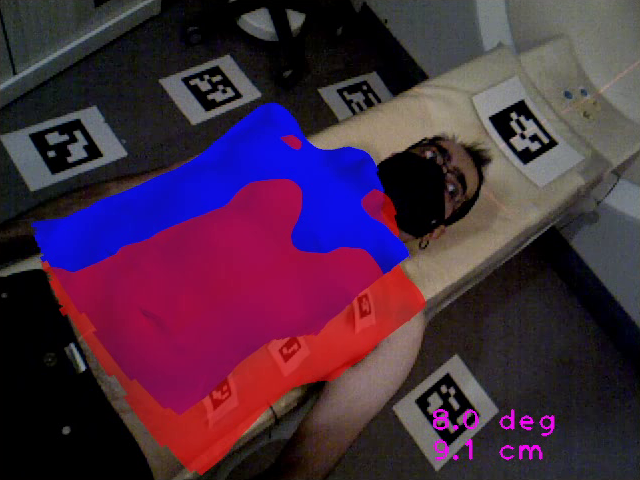}
    \includegraphics[width=.3\textwidth]{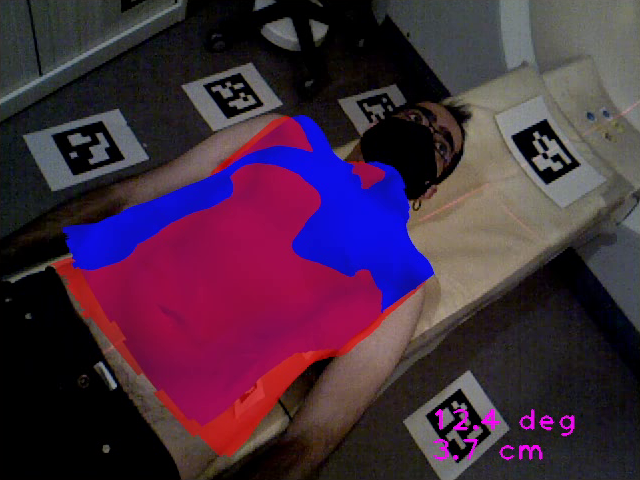}\\
    \includegraphics[width=.3\textwidth]{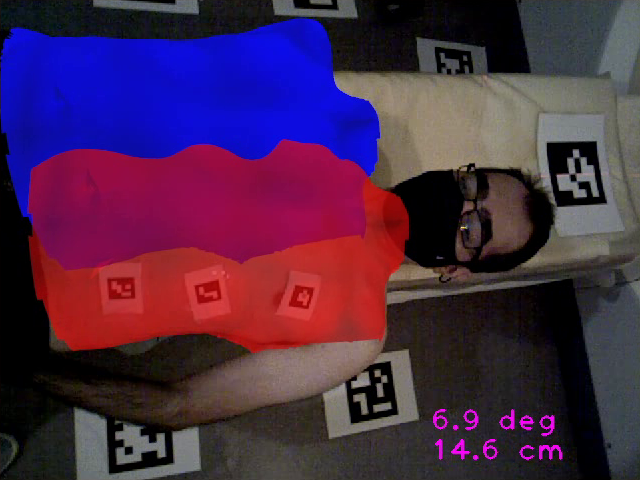}
    \includegraphics[width=.3\textwidth]{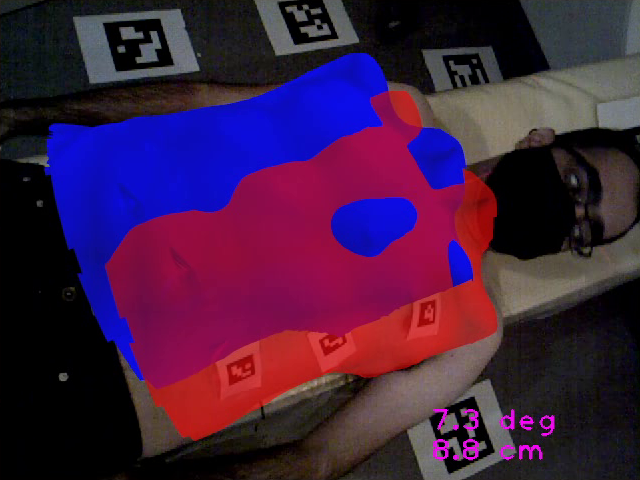}
    \includegraphics[width=.3\textwidth]{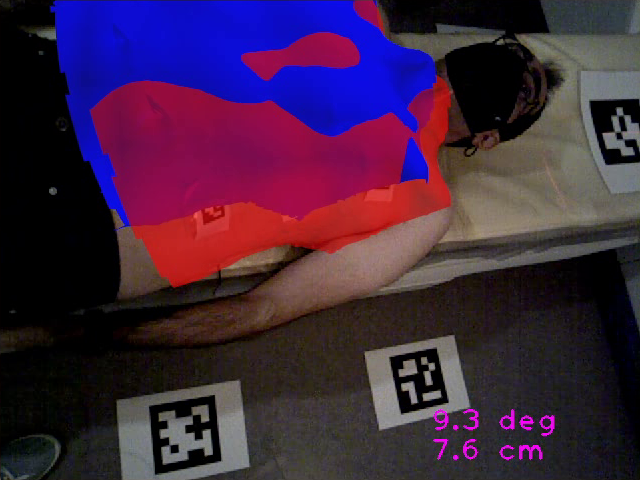}\\
    \includegraphics[width=.3\textwidth]{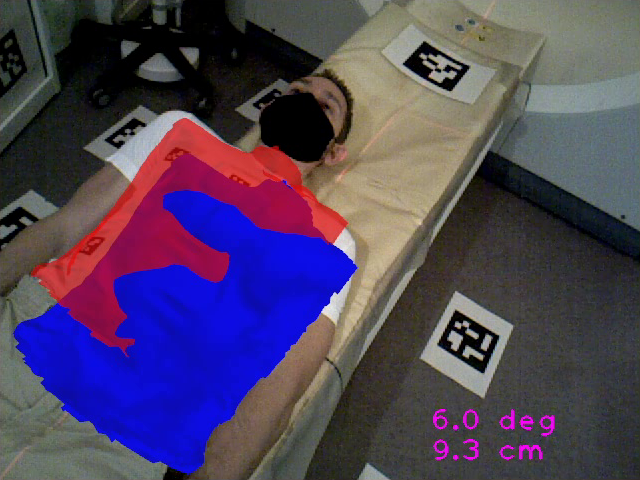}
    \includegraphics[width=.3\textwidth]{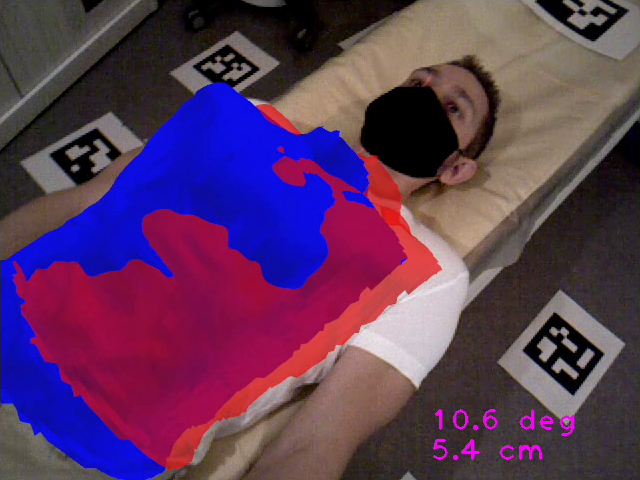}
    \includegraphics[width=.3\textwidth]{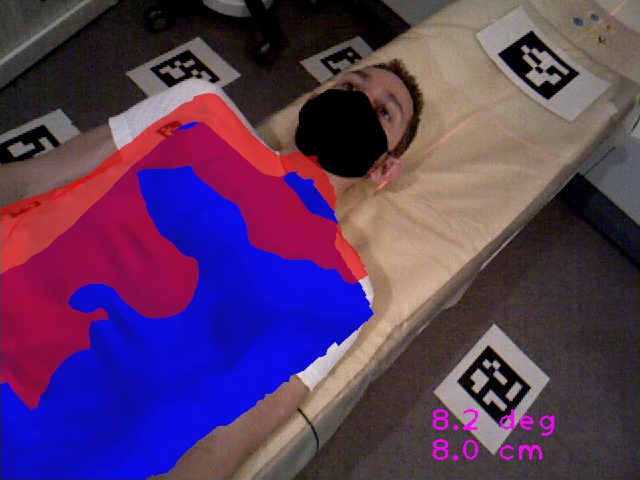}\\
    \includegraphics[width=.3\textwidth]{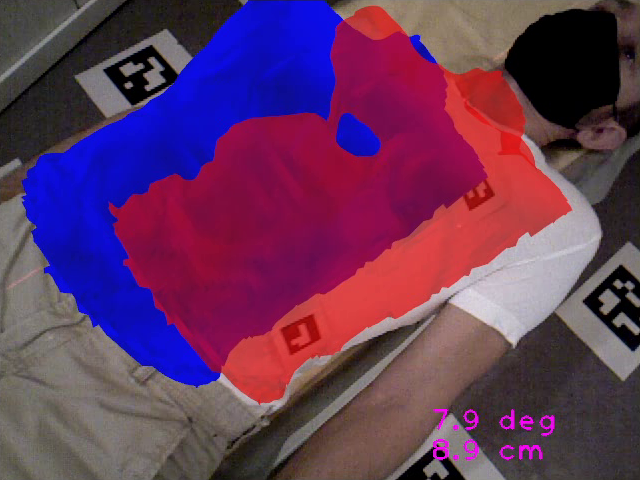}
    \includegraphics[width=.3\textwidth]{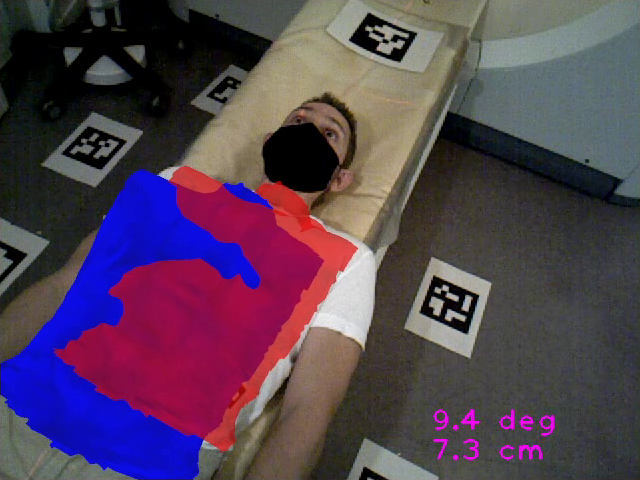}
    \includegraphics[width=.3\textwidth]{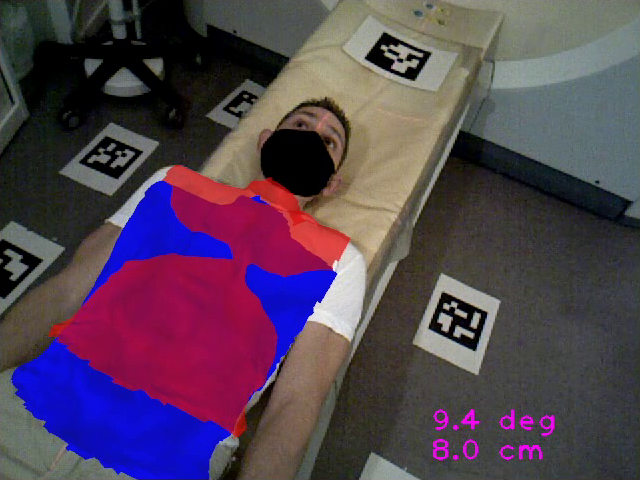}
    
    \caption{An example of the result from our tracking algorithm for patient positioning for multiple frames in different sequences for three human subjects. The red mesh shows the currently tracked patient pose, the blue mesh shows the pose where the patient needs to be positioned. There is a feedback of the pose error in rotation and translation in the corner of each frame.}
    \label{fig:tracking_demo}
\end{figure*}

\begin{figure*}[t!]
    \centering
    \includegraphics[width=.3\textwidth]{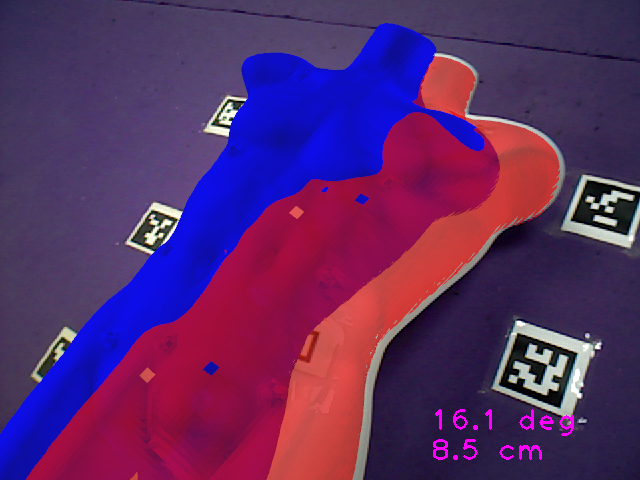}
    \includegraphics[width=.3\textwidth]{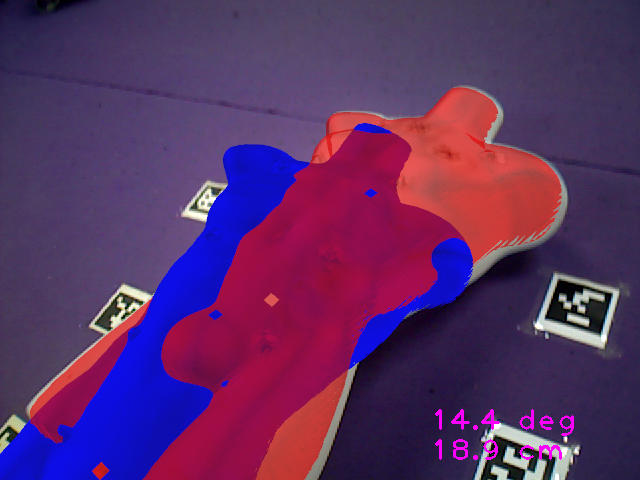}
    \includegraphics[width=.3\textwidth]{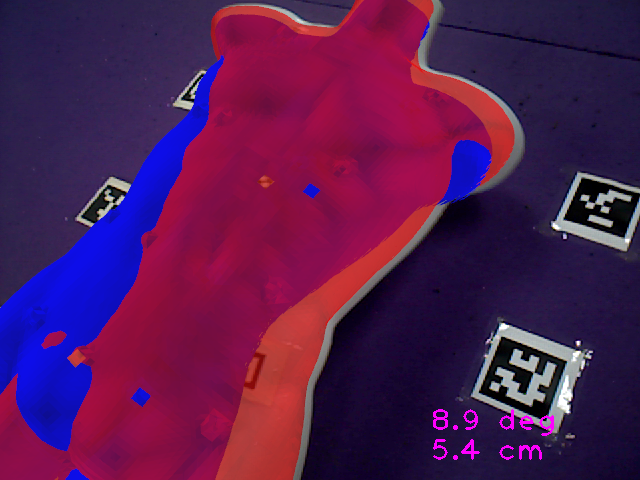} \\
    \includegraphics[width=.3\textwidth]{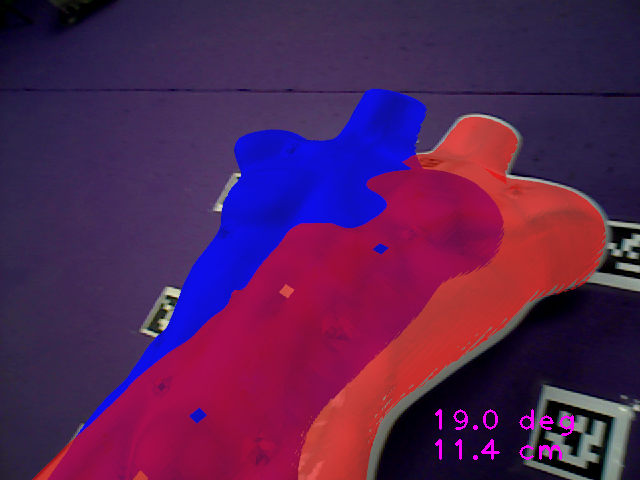}
    \includegraphics[width=.3\textwidth]{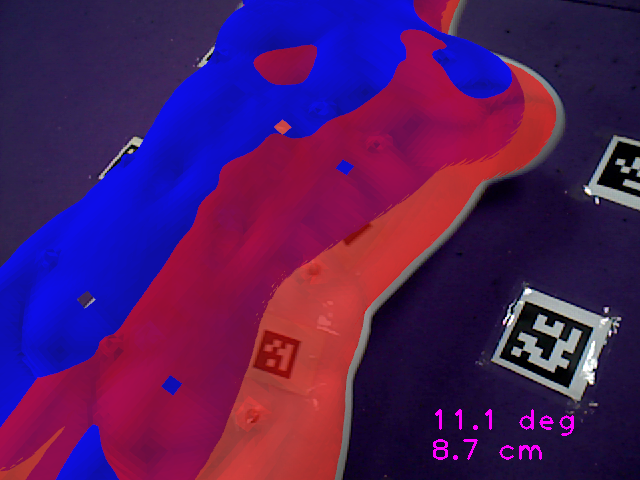}
    \includegraphics[width=.3\textwidth]{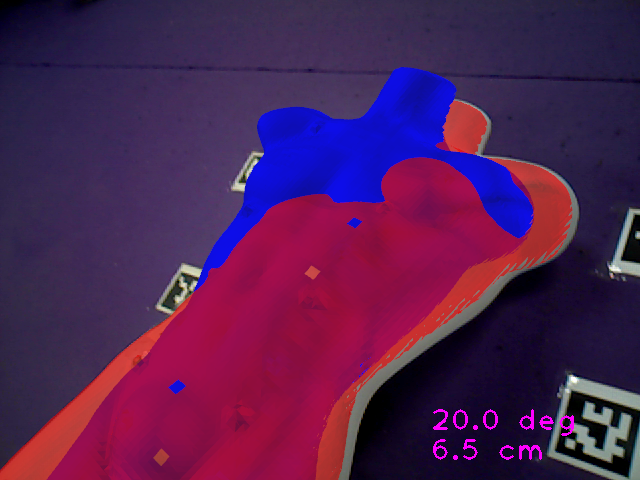}\\
    \caption{Qualitative results using the same mannequin employed in quantitative evaluation. The red mesh shows the current pose of the mannequin, and the blue mesh shows the desired pose. Rotational and translation error feedback is visualized in the corner of the images.}
    \label{fig:mannequin}
\end{figure*}

For qualitative results, we have applied our algorithm to several sequences on human subjects using the Xtion Pro Live sensor and also on the mannequin we have used in our quantitative evaluation.

We employed 3 different human subjects, one woman and two men presented in Figure \ref{fig:tracking_demo}. One man (middle two rows) has no apparel on his torso and the ArUco markers are attached on his skin. The second man (last two rows) has tight-fitting clothing on his torso  and the markers are attached on the clothing. For the woman (first two rows), the upper torso is clothed and the lower half of the torso is exposed with most of the markers attached to this part. In the figure, the red transparent model shows their tracked pose and the blue solid model shows their desired pose.

Finally, the snapshots related to the mannequin's qualitative evaluation are presented in Figure \ref{fig:mannequin}. The color codes are the same as the ones in Figure \ref{fig:tracking_demo} and similarly there is a numerical feedback of the positioning error. 

\section{Discussion}
\label{sec:discussion}
\subsection{Quantitative Results}
As it can be viewed in Table \ref{tab:quantitative} we where able to reach the average error value of \SI{7.28}{\milli\meter}/\SI{1.77}{\degree} using the Xtion Pro Live sensor and \SI{4.17}{\milli\meter}/\SI{0.82}{\degree} employing the Realsense L515 sensor. We suspect the main reason that we get a higher accuracy using the Realsense sensor is that it has a higher color image resolution. The higher resolution leads to a better ArUco marker detection and pose estimation which are employed both for scene tracking and patient tracking. It also helps with more robust detection of image features which is used by UcoSLAM for scene tracking.

Additionally the overall median errors (\SI{6.71}{\milli\meter}/\SI{1.53}{\degree} for the Xtion Pro Live sensor and \SI{3.83}{\milli\meter}/\SI{0.77}{\degree} for the Realsense L515 sensor) are significantly smaller than the average error which means that most of the errors come from the minority of frames. This suggests the general robustness of our method in pose estimation.

In regards to the tracking speed, the average frame rate of our implementation was 19 fps for the Xtion Pro Live sensor and 9 fps for the Realsense L515 sensor. We would like to suggest that the reason the slower frame rate for the Realsense sensor is because of its higher resolution color image (1280$\times$720). The higher resolution causes slower marker and image feature detection that affects both the scene tracking and the patient tracking speed. It should be reminded that this frame-rate was measured on a relatively old laptop. Hence we believe that our algorithm has the potential for real-time operation using up-to-date hardware. 

Regarding the running time of the preparation steps, as can be observed in Table \ref{tab:preparation_time}, we were able to prepare the program to start tracking in under 2 minutes for the Xtion Pro live sensor and under 3 minutes using the Realsense L515 sensor. Furthermore, it can be seen that the semi-automatic reference model alignment can be done relatively fast and the majority of the time needed by the preparation steps is spent on body model's map creation and its geometrical reconstruction.

\subsection{Qualitative Results}
 As you can see in Figure \ref{fig:tracking_demo} there are reliable pose estimations for all of the subjects. In some cases, you might notice small mismatches in the tracked torso (red model) and the actual current pose of the patient. We believe this is because of the non-rigid deformations of the upper body. Since we are not using any apparatus to fix the upper body it can slightly deform and cause a small error in registration. As said before, this can be improved by employing a fixing apparatus.
 
 In Figure \ref{fig:tracking_demo} you can also view our two ways of visual feedback. First, the model meshes related to the current pose and the target pose of the patient. Second, the numerical feedback on the bottom corner of the image showing the positioning error in rotation and translation. As you can see the intersection of the two model meshes can clearly show the operator how close the current pose of the patient matches that of the target pose. Also, when the patient pose is close enough to the target pose the operator can correct it furthermore by looking at the numerical position error visualized on the corner of the image.

Finally regarding the qualitative results of the mannequin, as can be observed in Figure \ref{fig:mannequin}, tracking is done with high accuracy and the tracked model matches almost perfectly with the mannequin. We believe that since there are no non-rigid movements here unlike the human subject and there is a smaller room for error. 

\section{Conclusion and future work}
\label{sec:conclusion}

This paper has proposed a cost-effective and efficient method for interactive AR in patient positioning that  can be used on mobile devices (such as Head Mounted Displays) requiring a only a RGBD camera and a regular CPU. Our method combines 3D information with texture information (keypoints) and artificial markers in order to create a map of the environment and a 3D model of the patient. In the planning phase, a map of the treatment room is created and the 3D model of the patient is obtained. Then, for each treatment session, a temporary body model is created at the beginning of the session, which is employed for tracking. Our system is able to simultaneously track the camera pose in the treatment room as well as the subject's body, providing visual information about the target position required for treatment.

The conducted experiments prove that our proposal achieves a high accuracy with the mean error of \SI{4.17}{\milli\meter}/\SI{0.82}{\degree} and the median error of \SI{3.83}{\milli\meter}/\SI{0.77}{\degree}. Also, the proposed method has proved to obtain a relatively high frame rate (9 fps) without the need to use a dedicated GPU nor a very powerful computer. It is then a cost-effective method that can be even employed in hospitals with a limited budget. Our qualitative results also showed the usefulness of the algorithm to be employed with the AR interface and how it can help the operator for patient positioning.

We still think that there is room for improvement. Our algorithm only performs rigid tracking similar to some other industrial level solutions. Adding non-rigid tracking the algorithm can make our approach even more capable for example in tracking the respiration movements of the patient. Another interesting future work is to accounting for local deviation of body parts when registering the model, since it could improve the system precision. We also think the application of our algorithm in other medical areas such as medical education can be explored further.

We also consider that it will be necessary in the future to conduct a study to obtain feedback from medical doctors on how our solution can be improved to be applied in real life situation.

Finally, we would like to mention that we have opened the way for other researchers to work on mobile interactive AR to assist patient positioning. We have laid down a framework that could be improved or build upon by others to have even better affordable patient positioning solutions that take advantage of AR.

\section*{Conflict of interest}

The authors do not have financial and personal relationships with other people or organizations that could inappropriately influence (bias) their work.

\section*{Acknowledgment}
  This project has been funded under projects TIN2019-75279-P and IFI16/00033 (ISCIII) of Spain Ministry of Economy, Industry and Competitiveness, and FEDER. The authors thank the Health Time Radiology Company for its help for the Project IFI16/00033 (ISCIII)

\bibliographystyle{plain}
\bibliography{references}
\end{document}